\documentclass[11pt]{article}
\usepackage[capitalise,noabbrev]{cleveref}
\usepackage{graphicx}
\usepackage{tabu}
\usepackage{amsmath}

\usepackage{acl2017}
\usepackage[normalem]{ulem}
\usepackage{textcomp}
\usepackage{amssymb}
\usepackage{microtype}
\usepackage{booktabs}
\usepackage{siunitx}
\usepackage{amsmath}
\usepackage{xspace}
\usepackage{tikz}
\usepackage{relsize}
\usepackage{url}
\usepackage{enumitem}
\usetikzlibrary{positioning}
\usetikzlibrary{bending}
\usetikzlibrary{arrows}
\usetikzlibrary{tikzmark}

\newenvironment{example}{\quotation}{\endquotation}
\usepackage{environ}
\makeatletter
\newsavebox{\measure@tikzpicture}
\NewEnviron{scaletikzpicturetowidth}[1]{%
  \def\tikz@width{#1}%
  \begin{lrbox}{\measure@tikzpicture}%
  \BODY
  \end{lrbox}%
  \pgfmathparse{#1/\wd\measure@tikzpicture}%
  \BODY
}
\makeatother

\aclfinalcopy 
\author{Hannes Schulz\thanks{\ \ Both authors contributed equally.}
  \and
  Jeremie Zumer\footnotemark[1]
  \and
  Layla El Asri
  \and
  Shikhar Sharma\\
  Microsoft Maluuba\\
\url{first.last@microsoft.com}\vspace{-5mm}}
\date{\today}
\title{A Frame Tracking Model\\for Memory-Enhanced Dialogue Systems}
 
 \DeclareMathOperator{\softmax}{softmax}
 
\begin{document}

\maketitle

\begin{abstract}
Recently, resources and tasks were proposed to go beyond state
tracking in dialogue systems. An example is the frame tracking task, which requires recording multiple
frames, one for each user goal set during the dialogue. This allows a user, for
instance, to compare items corresponding to different goals. This
paper proposes a model which takes as input the list of frames created so far
during the dialogue, the current user utterance as well as the dialogue acts,
slot types, and slot values associated with this utterance. The model then
outputs the frame being referenced by each 
triple of dialogue act, slot type, and slot value. We show that on the recently
published Frames dataset, this model significantly outperforms a previously proposed
rule-based baseline. In addition, we propose an extensive analysis of the frame
tracking task by dividing it into sub-tasks and assessing their difficulty with
respect to our model. 
\end{abstract}

\section{Introduction}
\label{sec:intro}

Conversational agents can seamlessly integrate into our lives by offering a natural language interface for complex tasks. 
However, the complexity of conversations with current slot-filling dialogue systems is limited.
One limitation is that the user usually cannot refer back to an earlier state in the dialogue, which is essential \textit{e.g.}, when comparing alternatives or researching a complex subject.

The recently published Frames dataset \citep{el_asri_frames_2017} provides 1369 goal-oriented human-human dialogues where the participants had to decide on a vacation package to purchase. The authors observed that in order to make up their minds, participants often compared different packages and referred to items that had been previously discussed during the dialogue. Current dialogue systems do not model the dialogue history in a way that a user can go back-and-forth between the different things that have been discussed. To address this shortcoming, \citet{el_asri_frames_2017} introduced a new task called \textit{frame tracking}. Frame tracking is an extension of the state tracking \citep{henderson_machine_2015,Williams:16b} task.

In a task-oriented dialogue system, the state tracker keeps track of the user goal. The user goal is often represented as the set of constraints that the user has (\textit{e.g.}, a budget) as well as the questions that the user has about the items presented to her by the dialogue system (\textit{e.g.}, the price of the vacation package). It is assumed that the dialogue system only needs to keep track of the last set of constraints given by the user. As a consequence, the user can change her goal during the dialogue but never come back to a previous goal. Frame tracking consists of recording all the different goals set by the user during the dialogue. This requires creating a new frame for each new user goal, which is the annotation provided with the Frames corpus. 

A frame tracker needs to be able to assign each new user utterance to the frames it references. This requires understanding which frame the user is talking about and recognizing when the user changes her goal, which implies that a new frame is created. For \textit{e.g.} comparisons, multiple referenced frames need to be identified. This paper proposes a neural model that attempts to solve these tasks.

We show that the model significantly outperforms the baseline proposed by \citeauthor{el_asri_frames_2017} on all the tasks required to perform frame tracking except for when the user switches frames without specifying slots. We also provide an analysis of frame tracking. In particular, we show that our model knows what frame anaphora refer to almost 90\% of the time, and which hotel is being talked about 84.6\% of the time. On the other hand, it does not perform well on slots which tend to be repeated in many frames, such as \texttt{dst\_city} (destination city). It also has difficulties selecting the right frame among similar \texttt{offer}s introduced in the same dialogue turn.

\section{Frame Tracking: An Extension of State Tracking}
\label{sec:frametracking}

\begin{figure}
\centering
\includegraphics[width=\linewidth]{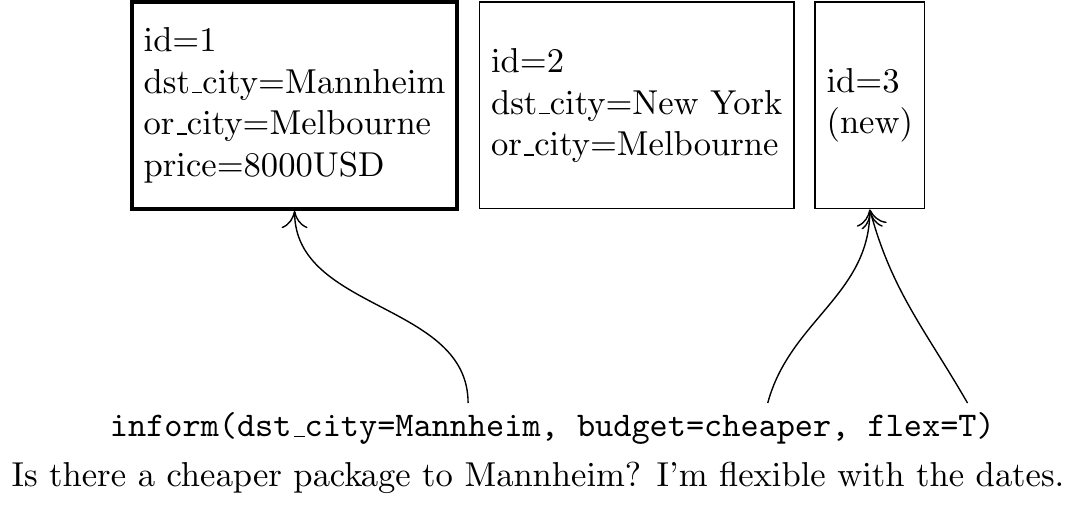}
\caption{Illustration of the frame tracking task. The model must choose, for each slot, which frame it is referring to, given the set of available frames, the previous active frame (bold), and the potential new frame (marked ``(new)'').} 
\label{fig:frametracking}
\end{figure}
In a goal-oriented dialogue system, the state tracker records the user goal in a semantic frame \citep{singh_optimizing_2002,raux_lets_2003,el_asri_nastia:_2014,laroche_final_2011}. The Dialogue State Tracking Challenge (DSTC) \citep{Williams:16b} defines this semantic frame with the following components:
\begin{itemize}[noitemsep]
\item User constraints: slots which have been set to a particular value by the user.
\item User requests: slots whose values the user wants to know.
\item User's search method: the user's way of searching the database (\textit{e.g.}, by constraints or alternatives).
\end{itemize}

In state tracking, when a new user constraint is set, it overwrites the previous one in the frame. In frame tracking, a new user constraint creates a new frame and thus, there are as many frames as user goals explored during the dialogue. 

To deal with user goals, two components specific to this setting were added to a frame, namely:
\begin{itemize}[noitemsep]
\item User's comparison requests: user requests for this frame and one or more other frames.
\item User's binary questions: user questions with slot types and slot values for this frame and possibly one or more other frames.
\end{itemize}
The user's search method is not part of the semantic frame defined for frame tracking. In addition, a frame is also created when the wizard proposes a vacation package to the user. This type of frame contains the same components as the ones defining a user goal except that the constraints are set by the wizard and not the user. A new frame is created so that if the user wants to consider the package, it is possible to switch to this frame and consider it to be the current user goal.

An example of comparison request is ``\textit{Could you tell me which of these resorts offers free wifi?}'' and an example of binary question is ``\textit{Is this hotel in the downtown area of the city?}'' or ``\textit{Is the this trip cheaper than the previous one?}''. A user request only has a slot type, \textit{e.g.}, ``\textit{Where is this hotel?}'' whereas a binary question has a slot type and a slot value. In other words, a binary question amounts to a confirmation and a request, to an open question.

Frames was collected using a Wizard-of-Oz method (WOz, \citealp{kelley_iterative_1984,rieser_corpus_2005,wen_network-based_2016}): for each dialogue a user and a wizard were paired. The user had a set of constraints and was tasked with finding a good vacation package that fits these constraints. The wizard had access to a database of vacation packages and helped the user find a suitable package. The wizards were thus playing the role of the dialogue system.

Each dialogue turn in the dataset is annotated with the currently active frame, \textit{i.e.}, the frame that is being currently discussed. The corpus was annotated in such a way that both users and wizards could create new frames. On the wizard side, a new frame is created whenever the wizard proposes a new package to the user. However, only the user can switch the currently active frame, for instance, by asking questions about a package proposed by the wizard. The motivation is that the user should have control of the frame being discussed throughout the dialogue, the dialogue system being an assistant to the user.

\citet{el_asri_frames_2017} define the frame tracking task as follows:
\begin{example}
For each user turn $\tau$, the full dialogue history $H = \{F_1,...,F_{n_{\tau-1}}\}$ is available, where $F_i$ is a frame and $n_{\tau-1}$ is the number of frames created up to that turn. The following labels are known for the user utterance $u_\tau$ at time $\tau$: dialogue acts, slot types, and slot values. The task is to predict if a new frame is created and to predict the frame or frames that are referenced in each dialogue act. A referenced frame can be the currently active frame or a previous one.
\end{example}
This task is illustrated in \cref{fig:frametracking}. We propose a model that tries to solve this task and analyze this model's behavior on several sub-tasks.




\section{Related Work}
As discussed in the previous section, frame tracking extends state tracking from only tracking the current user goal to tracking all the user goals that occur during the dialogue.

Recently, several approaches to state tracking have attempted to model more complex behaviors than sequential slot-filling. The closest approach to ours is the Task Lineage-based Dialog State Tracking (TL-DST) setting proposed by \citet{Lee:16}. TL-DST is a framework that allows keeping track of multiple tasks across different domains. Similarly to frame tracking, \citeauthor{Lee:16} propose to learn a dynamic structure of the dialogue composed of several frames corresponding to different tasks. TL-DST encompasses several sub-tasks among which \textit{task frame parsing} which consists of assigning a set of new dialogue acts to frames. This relates to frame tracking except that they impose constraints on how a dialogue act can be assigned to a frame and a dialogue act can only reference one frame. \citet{Lee:16} trained their tracking model on datasets released for DSTC (DSTC2 and DSTC3, \citealp{Henderson:14,Henderson:14c}) because no appropriate data for the task was available at the time. With this data, they could artificially mix different tasks within one dialogue, \textit{e.g.}, looking for a restaurant and looking for a pub, but they could not study human behavior and how humans switch between tasks and frames. Besides, TL-DST allows switching between different tasks but does not allow comparisons which is an important aspect of frame tracking. 

Another related approach was proposed by \citet{Perez:16}, who re-interpreted the state tracking task as a question-answering task. Their state tracker is based on a memory network \citep{weston:14} and can answer questions about the user goal at the end of the dialogue. They also propose adding other skills such as keeping a list of the constraints expressed by the user during the dialogue. This work did not attempt to formalize the different constraints as separate states to record.

Before describing our frame tracking model, we analyze the frame-switching and frame-creation behavior in Frames.

\section{Analysis of Frame References}
\label{sec:frames-analysis}
\subsection{Reasons for Referencing Other Frames}
\label{sec:frames-reasons}
The Frames dataset contains 19986 turns, among which 10407 are user turns. In 3785 (36\%) of these user turns, the active frame is changed. When the active frame is not changed, the user refers to one or more other frames in 7.5\% of the turns.

If we consider only \texttt{inform} acts\footnote{Utterances where the user informs the wizard of constraints.}, a dialogue system with a traditional
state tracker which tracks only a single semantic frame would be able to deal
with the subset of frame changes which correspond to overriding an already
established value (1684 turns or 44\% of the turns where frame changes occur).
The remaining 2102 (56\%) turns contain \texttt{switch\_frame} acts from the
user. The \texttt{switch\_frame} act indicates when a user switches from the
currently active frame to a previously-defined frame. A \texttt{switch\_frame}
act directly follows one or several vacation-package offers from the wizard in
1315 (38\%) of the frame-changing turns. In 428 turns, the user selects between
multiple offers made by the wizard and in 887 turns, she accepts a single offer
made by the wizard. A total of 787 (20\%) frame switches are made to a point in
the dialogue which is anterior to the directly preceding turn. Note that in this
work, we assume that we know the list of all previous frames at each turn of the
dialogue but a practical dialogue system should generate this list dynamically
during the dialogue. For this reason, it is crucial to also correctly interpret the
user's \texttt{inform} acts so that if a user appeals to an old frame, this
frame exists and is correctly identified.

Most of the turns where the user does not change the active frame but refers to other frames contain \texttt{request\_compare} (asking to compare different frames, 191), \texttt{negate} (98), \texttt{request} (28), and \texttt{request\_alts} (asking for another package, 17) acts.


%
%


\subsection{Examples}
\label{sec:examples}

In this section, we categorize instances of interesting frame-related user behavior and discuss the resulting requirements for a frame tracker.

\begin{itemize}[noitemsep]
\item \textbf{Switching to a frame by mentioning a slot value.} ``Oh, the Rome deal sounds much better!'', ``Can you tell me more about the Frankfurt package?'', ``I'll take the 13 day trip then!''.

For this case, we need to find which frames match the identified slot values, for instance, the destination city in the first example. Since there might be multiple matching frames, we have to incorporate recency information as well. 
In addition, equivalences have to be taken into account (13 --
thirteen, September -- sept, NY -- Big Apple, etc.).
Furthermore, in some cases, we need to learn equivalences between slots. \textit{E.g.}, the user has a budget, but the wizard typically only mentions prices.

\item \textbf{Switching to a frame without referencing it directly, usually by accepting an offer explicitly or implicitly.} ``yeah tell me more!'', ``yes please'', ``Reasonable. any free wifi for the kids?''.

The difficult part here is to identify whether the user actually accepted an offer at all, which also modifies the frame if the user asks follow-up questions in the same turn like in the third example. Some users ignore irrelevant wizard offers completely.

\item \textbf{Switching to a frame using anaphora.}
``Yeah, how much does the second trip cost?'', ``When is this trip and what is the price?'', ``Give me the first option, thank you''.

This is a slightly more explicit version of the previous case, and requires additional logic to determine the referenced frame based on recency and other mentioned slot values.

\item \textbf{Implicit reference for comparisons.}
``Do these packages have different departure dates?''.

\item \textbf{Explicit reference for comparisons.}
``Can you compare the price of this and the one to the package in St. Luis?'' (sic)

\item \textbf{Creating a new frame by specifying a conflicting slot value.}
``okaaay, how about to Tijuana then?'', ``what's the cheapest you got?'', ``Can I get a longer package if I opt for economy first?''

Here, the mentioned slot values need to be explicitly compared with the ones in the current frame to identify contradicting values. The same similarities discussed in frame switching above must be considered.
The context in which the slot values occur may be crucial to decide whether this is a switch to an old frame or the creation of a new one.

\item \textbf{Creating a new frame with an explicit reference to a previous one.}
``Are there flights from Vancouver leaving around the same time from
another departure city?'', ``I'd like to also compare the prices for a trip to Kobe between the same dates.'', ``Is there a shorter trip to NY?''.

In these examples, the slots time, date, and duration depend on references to frames (the current frame and the NY frame, respectively).
\end{itemize}

\section{Frame Tracking Model}
\label{sec:methods}

In the previous section, we identified various ways employed by the users to reference past frames or create new ones. In the following sections, we describe a model for frame tracking, \textit{i.e.}, a model which takes as input the history of past frames as well as the current user utterance and the associated dialogue acts, and which outputs the frames references for each dialogue act. 

\subsection{Input Encoding}
\label{sec:methods-in-enc}
Our model receives three kinds of inputs: the frames that were created before
the current turn, the current turn's user dialogue acts without frame
references, and the user's utterance. We encode these three inputs before
passing them to the network. The frames and the dialogue acts in particular are
complex data structures whose encodings are crucial for the model's performance.

\subsubsection{Text Encoding}
\label{sec:methods-text-enc}
We encode the user text as well as all the slot values by tokenizing the strings\footnote{using nltk's \texttt{TweetTokenizer}, \url{www.nltk.org}} and converting each token to letter trigrams\footnote{\textit{E.g.}, ``hello'' is converted to \#he, hel, ell, llo, lo\#}. Each trigram $t\in\mathcal T$ is represented as its index in a trainable trigram dictionary \(D_{\mathcal T}\).

\subsubsection{Frame Encoding}
\label{sec:methods-frame-enc}
We encode only the constraints stored in the set \(\mathcal F\) containing the frames created before the current turn.
In the Frames dataset, each frame \(F\in\mathcal F\) contains constraints composed of slot-value pairs, where for one slot \(s\in\mathcal S\) multiple equivalent values (\textit{e.g.}, NY and New York) and additional negated values (for instance if the user says that she does not want to go to a city proposed by the wizard) may be present. We encode a string representation of the most recent non-negated value \(v\) as described in \cref{sec:methods-text-enc}. The slot type is encoded as an index in a slot type dictionary \(D_{\mathcal S}\). 
The final frame encoding is the concatenation of all slot-value pairs in the frame.

In addition to the encoded frames, we also provide two vectors to the model: a one-hot code \(f_{c}\) marking the frame that was active in the last turn (the bold frame in \cref{fig:frametracking}) and a one-hot code \(f_{n}\) marking the frame that will be added if a new frame is created by the user in this turn (the frame marked ``(new)'' in \cref{fig:frametracking}).

\subsubsection{Similarity Encoding}
\label{sec:methods-similarity-enc}
To simplify learning of plain value matching, we precompute a matrix $S_L\in \mathbb{R}^{N{\times}\mathcal F}$, which contains the normalized string edit distance of the slot values in the user act to the value of the same slot in each frame, if present.

\subsubsection{Recency Encoding}
We also provide the model with information about the history of the dialogue by
marking recently added as well as recently active frames, coded as \(h^\tau_{d}\)
and \(h^\tau_{c}\), respectively, at turn \(\tau\). For a frame \(f\) introduced or last active at turn $\tau_f$, we set
\begin{align*}
\label{eq:recent_discount}
h_\cdot^\tau(f) = \begin{cases}
0 & \text{~if~} \tau < \tau_f\\
1 & \text{~if~} \tau = \tau_f\\
\gamma h_\cdot^{\tau-1} & \textnormal{otherwise.}\\
\end{cases}
\end{align*}


\subsubsection{Act Encoding}
\label{sec:methods-act-enc}

A dialogue act in the current turn has an act name \(a\in\mathcal A\) and a number of arguments.
Each argument has a slot type \(s\in\mathcal S\) and an optional slot value \(v\).
We use a dictionary \(D_{\mathcal A}\) to assign a unique index to each act \(a\), and use the same method as described in \cref{sec:methods-frame-enc} to encode slot-value pairs. 
In addition to the $N$ triples \((a,s,v)\), we encode every act
\(a\) separately, since an act may not have any arguments but still refer to a frame (cf. frame switching examples in \cref{sec:examples}).

\subsection{Output Encoding}
\label{sec:methods-out-enc}

For each triple \((a,s,v)\), our model outputs a multinomial distribution $p_{asv,F}$ over the frames \(F\in\mathcal F\).
Additionally, for each act \(a\in\mathcal A\) and frame \(F\in\mathcal F\), we determine the probability $p_{a,F}$ that $F$ is referenced by \(a\).

It can be difficult for the model to correctly predict the cases when the referenced frame is the currently active frame, especially in situations where (a) the slot values do not match and (b) the active frame was changed by an earlier act within the same turn.
To address this challenge, in the target, we replace all occurrences of the active frame with a special frame with index 0. In the example of \cref{fig:frametracking}, the value \texttt{flex=T} would point to this frame 0 since the active frame is changed by a previous value, in this case, the \texttt{budget}.

In the loss function, we do not penalize the model for confusing the active frame and the special frame except for \texttt{switch\_frame} and frame-creating \texttt{inform} acts, for which we want the model to predict the referenced frame. 

During prediction, we distribute $p_{asv,0}$ over $\mathcal F$ according to the predicted active frame:
\begin{align*}
 g_s & = \begin{cases}
 	1 & \text{if a \texttt{switch\_frame} act is present}\\
 	0 & \text{otherwise}
 \end{cases}\\
 p_{\text{new}} & = p_{asv,|\mathcal F|+1}\\
 p_{asv,F} &:= p_{asv,F} + 
              p_{asv,0} \big(
                  (1-g_s) \times p_{\text{new}} \\
                 & +g_s \times p_{\text{switch},F}\big),
\end{align*}
where $p_{\text{switch},\cdot}$ is the distribution assigned by the model to the \texttt{switch\_frame} act.
If no new frame was predicted and no \texttt{switch\_frame} act is present, the remaining probability mass is assigned to the previously active frame.

\begin{table*}[!t]
\begin{tabu}to\linewidth{@{}X[3,l]X[2,l]X[2,l]X[1.5,l]X[l]X[l]X[l]X[l]X[l]X[l]@{}}\toprule
    & \multicolumn{9}{c}{Accuracy (\%)}\\\cmidrule{2-10}
    Lesion & Full Acts & Only Acts & Frames & Text & $h_c^\tau$ & $h_d^\tau$ & $f_n$ & $S_L$ & $f_c$\\\midrule
    Slot-based & 58.3 & 66 & 63.7 & 74.5 & 65.4 & 78.8 & 79.5 & 64.4 & 82.7\\
    Act-based & 98 & 98.3 & 93.9 & 94.2 & 89.8 & 85.8 & 90.2 & 97.1 & 92.8\\
\bottomrule
\end{tabu}
\caption{Accuracy when removing model inputs.}
\label{tbl:lesion}
\end{table*}

\begin{figure}[tbp]
\centering
\includegraphics[width=\linewidth]{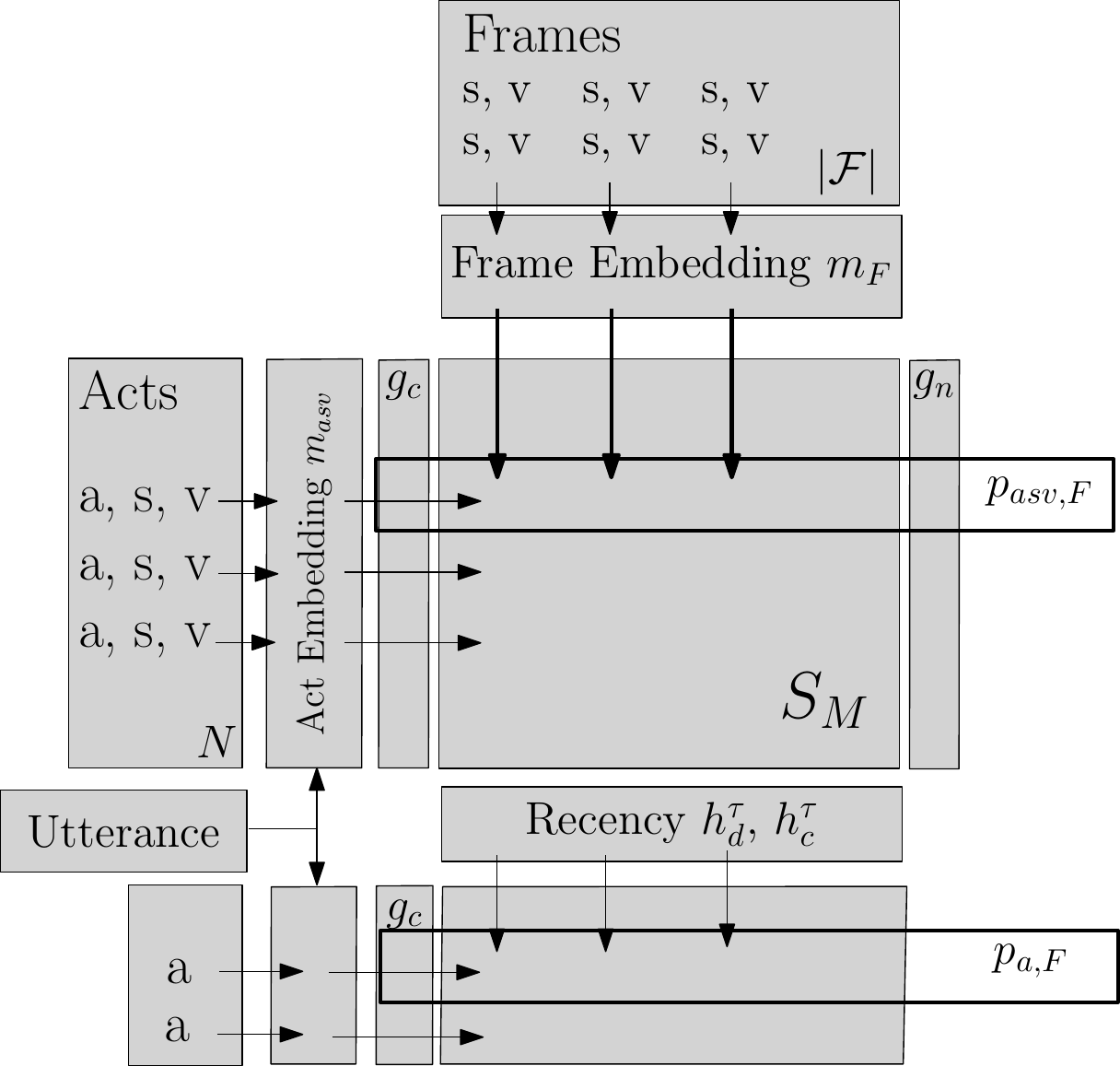}
\label{fig:schema}
\caption{Simplified overview of our model. $N$ triples of acts $a$ with slot-value arguments $s,v$ are matched to frames $F$ by computing a model similarity metric $S_M$. Frames are described by their constraints (slot-value pairs $s,v$). Together with the current and new frame indicators ($g_c$, $g_n$), $S_M$ represents a multinomial distribution $p_{asv,F}$ over the frames $F\in\mathcal F$. The same acts $a$ can refer to additional frames regardless of slot-value arguments, predicted in $p_{a,F}$ with the help of recency information $h^\tau$.}
\end{figure}

\subsection{Model Structure}
\label{sec:methods-model}

For each user turn, we first embed all dialogue acts \(a\), slot types \(s\), and letter trigrams \(t\) using the dictionaries \(D_{\mathcal A}\), \(D_{\mathcal S}\), and \(D_{\mathcal T}\), respectively. 
We sum the letter trigram embeddings for every token to generate trigram hashes
\citep{huang_learning_2013}. A bidirectional GRU \citep{cho_properties_2014} $r_t$ over
the hashes of values and the utterance generates summary vectors for both.
The summary vector is the concatenation of the final hidden state of the forward
and backward computation.

A second bi-directional GRU \(r_{asv}\) computes a hidden activation for each of the (act, slot, value) triples in the current turn.
We compute a value summary vector $m_{asv}$ by appending each hidden state of $r_{asv}$ with the utterance embedding and projecting to a 256-dimensional space.

For the frames, we proceed in a similar manner, except that the frames do not contain dialogue acts nor an utterance, so we use a GRU $r_F$ to compute hidden states for all slot-value pairs
\begin{align}
\left(\begin{matrix}
    D_{\mathcal S}[s_{1}], r_t(D_{\mathcal T}[v_{1}])\\ D_{\mathcal S}[s_{2}], r_t(D_{\mathcal T}[v_{2}])\\
    \ldots\end{matrix}\right).
\end{align}
During training, the constraint order within frames is shuffled.
The final hidden of the state $r_F$ is projected to a 256-dimensional space, resulting in a frame summary vector $m_F$.

By comparing slot values $m_{asv}$ mentioned by the user to the frames $m_F$, and
taking into account the recently-active and recently-added information, we can
determine which frame the user is referencing. To this end, we compute the
dot-product between $m_{asv}$ and $m_F$, resulting in a model similarity matrix
$S_M\in\mathbb R^{N{\times}|\mathcal F|}$. It is important to have the user
utterance in the value summary vector because without it, the comparison with
the frames would only work if slot values were explicitly mentioned, which is
not true in general. Boolean values, for example, are usually only
present implicitly (\textit{cf.} \cref{sec:examples}). 
We learn the weights of a linear combination of the model similarity matrix with the input $S_L$, yielding the final similarity matrix $S$.

Two special cases remain: (1) no match could be found and (2) a new frame should be created. To handle these cases, we extend $S$ with two columns corresponding to the active frame $g_c$ and the new frame $g_n$. Intuitively, $g_n$ is high if no frame matches the user turn and if there is a strong discrepancy with the active frame. On the other hand, $g_c$ is high only if no frame matches the user turn. Since again, the actual user utterance sometimes contains crucial information, we condition $g_n$ and $g_c$ on the maximum match with any frame, the match with the previously active frame, and the user utterance embedding.
 
For a user input triple \((a,s,v)\), the slot-based frame prediction is then computed as
\begin{align}
 p_{asv} = \softmax(&g_c, S_{asv,1}, S_{asv,2}, \ldots,\\
 	&\nonumber S_{asv,|\mathcal F|}, g_n).
\end{align}

Finally, we determine the act-based probability of frame references $p_{a,F}$.
For every pair $(a, F)$, this probability is computed by a 2 layer densely connected network conditioned on the dialogue act, the recency information, and the user utterance embedding. We also set $p_{a, 0}$ to $1-\max_F p_{a,F}$ to produce an implicit reference to the active frame by default.


\section{Experiments}
\label{sec:exp}
\subsection{Learning Protocol and Metrics}
We train the model by splitting the dataset into 10 folds as described by \citet{el_asri_frames_2017}. For each fold, we further split the training corpus into training and validation sets by withholding a random selection of 20\% of the dialogues from training. We use the Adam \citep{kingma_adam:_2014} algorithm to minimize the sum of the loss for $p_{asv}$ and $p_{a,F}$, with a learning rate of $10^{-3}$. Learning is stopped when the minimum validation error has not changed for ten epochs. We compare our model to the simple rule-based baseline described by \citet{el_asri_frames_2017}.

For slot-based predictions ($p_{asv,F}$), we report mean accuracy over the ten folds of the Frames dataset. For act-based predictions ($p_{a,F}$), \textit{i.e.}, we determine for every act $a$ whether the ground truth set of referenced frames is equal to the predicted set of referenced frames (with a cutoff at $p_{a,F}=\frac12$), and again average accuracy scores over the ten folds.

Results are summarized in \cref{tbl:folds_perf}. Our model strongly outperforms the baseline both on references with and without slots. In particular, we observe that our model excels at predicting frame references based on acts alone, while the baseline struggles to solve this task.

\subsection{Comparison with the Baseline}
\begin{table}[!t]
\begin{tabu}to\linewidth{@{}X[1.1,l]X[l]X[l]@{}}\toprule
    &\multicolumn{2}{c}{Accuracy (\%)}\\\cmidrule{2-3}
	& \textbf{Ours} & \textbf{Baseline}\\\cmidrule{2-3}
    Slot-based & \textbf{76.43}$\pm$4.49 & 61.32$\pm$2.19\\
    Act-based & \textbf{95.66}$\pm$2.34 & 66.81$\pm$2.58\\
\bottomrule
\end{tabu}
\caption{Performance comparison between the baseline of \citet{el_asri_frames_2017} and our model.}
\label{tbl:folds_perf}
\vspace{-0.3cm}
\end{table}

\begin{table}[!t]
\begin{tabu}to\linewidth{@{}X[3.0,l]X[0.5,r]X[1.0,r]@{}}\toprule
	&\multicolumn{2}{c}{Accuracy (\%)}\\\cmidrule{2-3}
    &\textbf{Ours} & \textbf{Baseline}\\\cmidrule{2-3}
    Frame change (new val) & 52.5 & 4.2\\
    No frame change (new val) & 93.8 & 74.3\\
    Frame change (no offer) & 36.4 & 22.7\\ 
    Frame change (offer) & 67 & 62.2\\
    request\_compare & 70.5 & 40.9\\
    \bottomrule
\end{tabu}
\caption{Partial comparison table of performance for different dialogue settings (cf \cref{sec:frames-analysis}), including frame changes/lack of frame changes upon the introduction of new values, as well as when preceded by an offer or not, demonstrating our model's improvements over the baseline.}
\label{tbl:event_types}
\end{table}

\begin{table}[!t]
\begin{tabu}to\linewidth{@{}X[2.7,l]X[0.7,r]X[1.3,r]@{}}\toprule
	&\multicolumn{2}{c}{Accuracy (\%)}\\\cmidrule{2-3}
    &\textbf{Ours} & \textbf{Baseline}\\\cmidrule{2-3}
    \texttt{switch\_frame(dst\_city)} & 66.1 & 21.4\\
    \texttt{switch\_frame(duration)} & 52.6 & 26.3\\
    \texttt{inform(seat)} & 60.0 & 36.0\\ 
    \texttt{request(end\_date)} & 66.7 & 0.0\\
    \bottomrule
\end{tabu}
\caption{Partial comparison table of act-slot combinations between our model and the baseline of \cite{el_asri_frames_2017}.}
\label{tbl:act_slot}
\end{table}
We further analyze the difference in performance between our frame tracking model and the rule-based baseline on classes of predictions on a single fold of the data. We organize the turns in the test set into 11 classes and measure performance by computing accuracy only on turns that fall into the respective class.

We first observe that the baseline model almost completely fails to identify frame changes when a new value is introduced by a user (4\% accuracy over 303 turns), frame changes associated with \texttt{switch\_frame} acts that do not have slot values, or when a \texttt{switch\_frame} act is present in a turn following one that does not contain an \texttt{offer} act. On the other hand, the baseline model predicts lacks of frame changes (74.3\% over 1111 instances) and frame changes after an \texttt{offer} (62.2\% over 312 instances) quite well.

Our model dominates the rule-based baseline on all classes except for the prediction of frame changes with \texttt{switch\_frame} acts that do not have slot values (4.2\% over 24 occurrences). Partial comparison results are presented in \cref{tbl:event_types}.

Perhaps surprisingly, our model correctly predicts 70.5\% of frames associated with \texttt{request\_compare} acts whereas the baseline only correctly identifies 40.9\% of them.

We then computed the accuracy on the set of unique act and slot combinations in
the dataset.
Here, our model outperforms the baseline on all act-slot pairs with more than 10 occurrences in the test set. We observe that the baseline performs quite poorly on \texttt{switch\_frame}s with \texttt{dst\_city} (destination city) slots, whereas our model does not have such a drawback. The same is true for a \texttt{switch\_frame} with a \texttt{duration} or for an \texttt{inform} with a \texttt{seat} (economy or business flight seat) or even a \texttt{request} with an \texttt{end\_date}. Results are presented in \cref{tbl:act_slot}. We note that our model performs worse on combinations that should express a match with a frame whose slot values use very different spellings (such as rich abbreviations and synonyms) whereas the baseline model is the weakest when slot values can be easily confused for values of other slots (\textit{e.g.} a rating of 5 (stars) vs. a duration of 5 (days)). Our model is also currently unable to distinguish between similar offers introduced in the same turn.

Code to generate the full set of metrics will be made available.

\subsection{Lesion Studies}
To assess which of the features are useful for the model, we remove the model's
inputs one at a time and measure the model's performance. Results are shown in 
\cref{tbl:lesion}. We observe that the model stops learning (\textit{i.e.} its
performance does not exceed the baseline's) on the act-slot-value triples when any of
the input is removed except for the new frame history, new frame candidate, and
previous frame inputs. Similarly, the model performance suffers when the new
frame candidate, any historical data, or the frames are removed. We observe
that all the inputs are used by the model in its predictions either for
$p_{asv,F}$ or $p_{a},F$.


\section{Discussion}
\vspace*{-2mm}
Our model makes use of the text to correctly predict the frames
associated with acts. Dependence on input text means our method is
domain-dependent. The annotation process for the Frames dataset is costly, so it
would be beneficial if we could transfer learned frame switching behavior to
other domains, possibly with already existing NLU components. A possible
solution might be to standardize the text after NLU, and use anonymous
placeholders instead of domain-specific words.

Additionally, our current model assumes a perfect NLU to provide acts, slots,
and values as inputs. While this is helpful for researching the frame
referencing issues in isolation, both components should work together. For
example, currently, we assume that a \texttt{switch\_frame} act is correctly
identified, but we do not know the frame the user wants to switch to. In a more
realistic pipeline, these decisions are closely related and also need to take
more of the dialogue history into account.

\subsection{Conclusion}

In this paper, we provided a thorough analysis of user behavior concerning
switching between different user goals in the Frames dataset. Based on this
analysis, we have designed a frame tracking model that outperforms the baseline
of \citet{el_asri_frames_2017} by almost 20\% relative performance. This model
assigns the dialogue acts of a new user utterance to the semantic frames created
during the dialogue, each frame corresponding to a goal. We analyzed the
strengths and weaknesses of the rule-based baseline and of our model on
different subtasks of frame tracking. Our model outperforms the baseline on all
but one subtasks. We showed that further improvement is necessary for
matching slot values when they are present in many distinct frames. We
have demonstrated that the frame tracking task can be performed effectively by
learning from data (our model correctly identifies frame changes in about 3 out
of 4 cases). This represents a first step toward memory-enhanced dialogue
systems which understand when a user refers to an older topic in a conversation
and which provide more accurate advice by understanding the full context of a
request.

\bibliography{joined}
\bibliographystyle{acl_natbib}
\end{document}